\pdfoutput=1

\documentclass[conference]{IEEEtran}
\IEEEoverridecommandlockouts

\usepackage{cite}
\usepackage{amsmath,amssymb,amsfonts}
\usepackage{graphicx}
\usepackage{booktabs}
\usepackage{xcolor}
\usepackage[hidelinks]{hyperref}

\newif\ifanon\anonfalse

\newif\ifdraft\drafttrue
\ifdraft
  \newcommand{\todo}[1]{\textcolor{red}{\small[\textbf{TODO:} #1]}}
\else
  \newcommand{\todo}[1]{\fbox{\textbf{UNRESOLVED}}}
\fi

\newcommand{\zt}{$z_t$}
\newcommand{\zhat}{$\hat{z}_{t+1}$}

\begin{document}

\title{Latent Telepathy: Multi-Robot Communication\\
with Self-Supervised Perceptual Latents}

\ifanon
  \author{\IEEEauthorblockN{Anonymous Submission}}
\else
  \author{%
    \IEEEauthorblockN{Howard Wang}
    \IEEEauthorblockA{\textit{Columbia University}\\
      New York, NY, USA\\
      hw3214@columbia.edu}
    \and
    \IEEEauthorblockN{Han Zheng}
    \IEEEauthorblockA{\textit{Laboratory for Information}\\
      \textit{and Decision Systems}\\
      \textit{Massachusetts Institute of Technology}\\
      Cambridge, MA, USA\\
      hanzheng@mit.edu}
    \and
    \IEEEauthorblockN{Cathy Wu}
    \IEEEauthorblockA{\textit{Laboratory for Information}\\
      \textit{and Decision Systems}\\
      \textit{Massachusetts Institute of Technology}\\
      Cambridge, MA, USA\\
      cathywu@mit.edu}%
  }
\fi

\maketitle

\begin{abstract}
In a decentralized multi-robot team under partial observability, the fact that decides a
robot's next action is often visible only to a teammate. Existing decentralized methods
communicate kinematic information, such as position or planned trajectory, which cannot
convey what the teammate perceives. Learned communication in multi-agent reinforcement
learning (MARL) can carry perceptual content, but the resulting messages are task-coupled
and opaque. We propose \emph{Latent Telepathy}. Each robot broadcasts the perceptual
latent vector it already computes for its own use, the output of an encoder trained with
a self-supervised joint-embedding predictive objective, frozen, and shared across the
team. A teammate learns to act on it from task reward alone. Because the encoder already
runs for perception, the message costs no additional computation and a single compact
vector of bandwidth. Because the encoder is frozen before any policy is trained, the
message means the same thing to every robot, and the receiving robot is never told what
it means. We evaluate Latent Telepathy with a content-controlled protocol in which
bandwidth, latency, topology and receiver are held fixed and only the message content
varies. Broadcasting the latent lets a navigator avoid an occluded hazard in $99.7\%$ of
episodes, matching a noiseless hand-designed message. Position and trajectory messages,
which cannot carry what the scout sees, remain at chance, and the raw camera image,
$186$ times wider, is less reliable than the compressed latent. The result holds from a
discrete gridworld to rendered pixels under continuous velocity control, and the encoder
decodes the hazard from a physical robot's camera in $102$ of $102$ live decisions. Along
the way we identify a requirement for porting MARL communication results to continuous
control, that the decision a message informs must remain reachable by exploration, and
show how to restore it.
\end{abstract}

\section{Introduction}

\begin{figure}[t]
\centering
\includegraphics[width=\columnwidth]{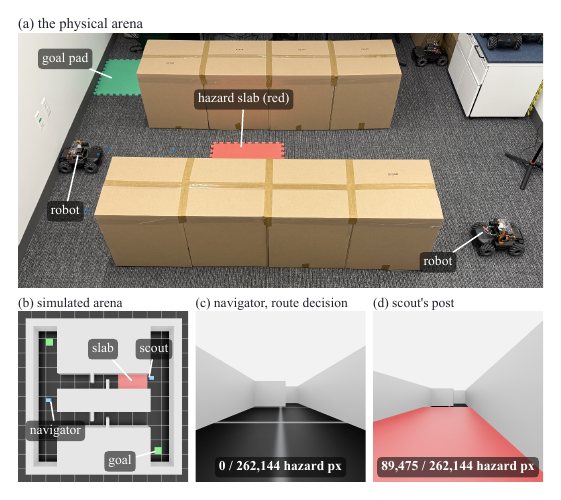}
\caption{Latent Telepathy: broadcast what you see. \textbf{(a)}~The physical arena of the hardware test
(Section~\ref{sec:hardware}): outer walls and a central divider forming one corridor each
side, the red hazard slab, both robots, and the goal pad. \textbf{(b)}~The simulated
arena: a barrier pierced by two corridors, a slab in one of them, a stationary scout
that can see it and a navigator that cannot. \textbf{(c)}~At the route decision the
navigator's camera contains $0$ of $262{,}144$ hazard pixels under either slab placement;
\textbf{(d)}~the scout's post contains $89{,}475$. The decisive bit exists only on the
scout's side of the channel. Broadcasting the scout's frozen self-supervised latent lets
the navigator choose the clear corridor in $99.7\%$ of episodes; broadcasting its
position or trajectory leaves the navigator at chance.}
\label{fig:scene}
\end{figure}

Consider two ground robots searching a building. One has turned a corner and can see
that the corridor ahead is blocked; the other, still behind the wall, is about to commit
to that corridor. Nothing in the second robot's own sensors can prevent the mistake.
This is the ordinary situation of a decentralized team under partial observability. Each
robot holds an egocentric, occluded view, and the decisive fact is frequently on the
other side of a wall. Centralized tracking with overhead cameras sidesteps the problem
but fails the moment a robot leaves the tracked volume or a packet drops. Decentralized
teams have to communicate, and the question we address is what a robot should send.

Today the answer is almost always kinematic. Intention-sharing methods broadcast a
robot's pose and a short planned trajectory~\cite{kim2021intention}; multi-agent
path-finding methods exchange positions and local plans~\cite{sartoretti2019primal,ma2021dhc}.
These messages are compact and interpretable, but a teammate's coordinates say nothing
about the obstacle behind it. The MARL literature on learned
communication~\cite{sukhbaatar2016commnet,das2019tarmac,singh2019ic3net} goes further
and trains the message end-to-end with the policy, so the message can in principle carry
perceptual content. In practice such messages are opaque, coupled to the task they were
trained on, and expensive to obtain, and they have mostly been demonstrated on
low-dimensional gridworlds where perception is not the bottleneck~\cite{liu2020who2com}.

Our starting observation is that every robot already computes a perceptual
representation for its own use. Encoders trained with joint-embedding predictive
objectives~\cite{assran2023ijepa,assran2025vjepa2} produce compact latent vectors that
capture the structure of a scene, without labels and without reward, and are increasingly used
as the perception front end for control~\cite{zhou2024dinowm}. To date these latents
stay on the robot that computed them. Latent Telepathy broadcasts them
(Fig.~\ref{fig:scene}). Three design choices follow. (i)~The encoder is
\emph{self-supervised and predictive}, so the latent keeps the geometric content a
teammate needs and drops the appearance detail it does not; that is what makes it a
small message. (ii)~The encoder is \emph{frozen before any policy exists and shared
across robots}, so every message means the same thing to every receiver, nothing drifts
with the task, and the same weights deploy without retraining. (iii)~The receiver is
\emph{trained by reinforcement learning from the team's task reward}, because in a
deployed team no one can label what a teammate should say; the only supervision
available is whether the task succeeded.

We evaluate Latent Telepathy with a content-controlled protocol. Bandwidth, one-step
latency, topology and receiver architecture are held fixed and only the message content
varies. Six contents are compared, from an empty message to the raw image, with a
noiseless oracle as a diagnostic. In a gridworld with engineered occlusion, the frozen latent solves an occluded-hazard
task in five of five seeds while position and trajectory messages stay at chance in
fifteen of fifteen. On rendered pixels under continuous velocity commands, it reaches
$99.7\%$ route optimality against a $49.6\%$ no-message floor, matches the oracle, and
outperforms the raw image at $1/186$ of the bandwidth. Corrupting the
message at evaluation returns the trained policies to chance. We also test the frozen
encoder on a physical robot's camera, where it decodes the hazard in $102$ of $102$ live
decisions.

We should be candid about one thing the port to continuous control taught us, because we
think it is the most transferable finding in the paper. The gridworld result transferred
to pixels immediately; it did not transfer to velocity control, and six training
configurations returned null results before we understood why. In a gridworld one action
moves the agent a whole cell, so choosing the other corridor is a single draw of the
policy and every training batch contains both routes. Under a Gaussian policy over
velocity commands the same choice requires roughly thirty consecutive steps of
coordinated lateral deviation, which independent per-step noise almost never produces. A decision that
is never sampled generates no advantage signal, so there is nothing for a message to be
recruited to inform, and even a noiseless ground-truth bit fails. Section~\ref{sec:port}
shows why exploring harder cannot fix this and how restoring the route as a single
temporally extended decision does.

The contributions of this paper are:
\begin{itemize}
\item We propose Latent Telepathy: frozen self-supervised perceptual latent vectors as
the communication primitive for decentralized robot teams, with a receiver trained from task
reward alone. To the best of our knowledge, this is the first demonstration that a
receiver can learn to act on such a message with no labels, no auxiliary loss and no
gradient into the encoder.
\item We introduce a content-controlled evaluation protocol with corruption controls that
isolates message content as the only variable, and show that the latent separates from
position and trajectory messages at matched bandwidth and matches a noiseless oracle.
\item We show that MARL communication results built on discrete actions do not transfer
to continuous velocity control without restoring temporal abstraction, give a scalar
predictor of when the failure occurs, validate it in $720$ independent runs, and give a
small architectural remedy.
\item We show on hardware that the frozen encoder's hazard content survives transfer to
a real camera.
\end{itemize}

\section{Overview}
\label{sec:overview}

\begin{figure}[t]
\centering
\includegraphics[width=0.95\columnwidth]{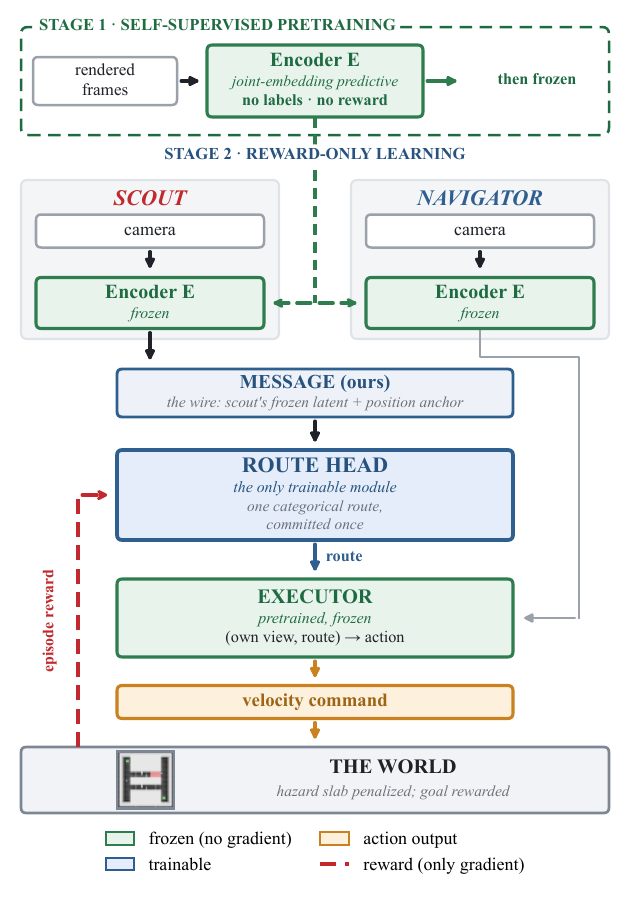}
\caption{The pipeline. \textbf{Stage 1} trains the encoder $E$ on rendered frames with a
joint-embedding predictive objective, using no labels and no reward, then freezes it;
both robots receive the same weights. \textbf{Stage 2} learns from reward alone. The
scout's latent and a two-float position anchor form the message; the only trainable
module is the part of the navigator that reads it, here a route head that commits to one
categorical route for a frozen executor to drive. The scalar episode reward (red) is the
only gradient, and it reaches the head alone.}
\label{fig:pipeline}
\end{figure}

We address the problem of a decentralized robot team acting under partial observability,
where the fact that decides one robot's next action is visible only to a teammate. Our
approach has three components (Fig.~\ref{fig:pipeline}). A perceptual encoder $E$,
shared by every robot, maps a camera image to a compact latent vector. A communication
step broadcasts the scout's latent \zt{} to the navigator at every control step,
prefixed with an anchor giving the scout's position relative to the receiver. A receiver
on the navigator pools the messages it gets with masked attention and turns the result
into an action. Under continuous control the receiver is a route head that commits to
one route per episode, and a low-level controller drives that route from the navigator's
own camera view.

The components are trained in two stages. In the first, $E$ is trained on frames from
random rollouts with a joint-embedding predictive objective, using no labels and no
reward, then frozen and copied to both robots. In the second, the receiver is trained by
reinforcement learning from the team's episode return. Nothing else is trained. The
encoder never sees reward, the low-level controller is pretrained and frozen, and the
navigator is never told what \zt{} means. In what follows we formalize the setting
(Section~\ref{sec:formulation}), describe each component (Section~\ref{sec:method}), and
explain why the route must be committed once per episode under continuous control
(Section~\ref{sec:port}).

\section{Problem Formulation}
\label{sec:formulation}

We consider a team of $N$ robots in a decentralized partially observable Markov decision
process with communication~\cite{oliehoek2016decpomdp}. At step $t$ robot $i$ receives an
egocentric observation $o^i_t$, emits a message $m^i_t = f(o^i_t) \in \mathbb{R}^D$ onto
a shared channel, and selects an action
$a^i_t \sim \pi_i(\,\cdot \mid o^i_t, \{m^j_{t-1}\}_{j \in \mathcal{N}(i)})$, where
$\mathcal{N}(i)$ is the set of robots within communication range and messages arrive
with one step of latency. The team receives a shared reward $r_t$, and $r_t$ is the
only training signal available to the policies. The message function $f$ is fixed
before training and identical across robots. The question this paper asks is which $f$
a robot should use, holding everything else about the channel constant.

We consider four kinds of $f$. A \emph{kinematic} message reports where the sender is or
where it intends to be, which is what deployed multi-robot systems send today. A
\emph{perceptual} message reports what the sender sees, either as the latent of its
current frame, $z_t = E(o^i_t)$, or as the latent it predicts for its next frame,
$\hat{z}_{t+1} = P(z_t, a^i_t)$, where $E$ is a self-supervised encoder and $P$ the predictor
trained alongside it. A \emph{raw} message sends $o^i_t$ itself. It is infeasible for a
radio link and serves as a ceiling on what the channel could carry. A \emph{diagnostic
oracle} sends the ground-truth bit the receiver needs. The oracle is
not a baseline. It is a noiseless hand-designed message that separates representation
failure from optimization failure, and we use it that way in Section~\ref{sec:port}.

For the comparison to be about content alone, everything else must be held fixed. Every
message is padded to a common width $D$, carries the same anchor
(Section~\ref{sec:method}), travels the same channel with the same latency and reaches
the same receiver trained under the same budget. We call this a content-controlled
protocol. Its instantiation for our task is given in Section~\ref{sec:setup}.

\section{Method}
\label{sec:method}

\subsection{Encoder and freeze}

The encoder $E$ is a small convolutional network that maps an egocentric frame to a
latent vector of a few dozen dimensions. It is trained with a joint-embedding predictive
objective. A predictor $P$ takes the latent of one frame and the action taken and is
asked to produce the latent of the next frame, as computed by an
exponential-moving-average target copy of $E$. VICReg variance and covariance terms guard against
collapse~\cite{bardes2022vicreg}, and a class-weighted spatial reconstruction auxiliary,
which we found necessary in the gridworld, is carried forward. Training data are frames
from random rollouts in the environment, with no labels and no reward. Before freezing,
the checkpoint must pass pre-registered gates. Linear probes confirm that the content a
teammate would need is decodable and an effective-rank check confirms that the latent
has not collapsed. The auxiliary heads are then discarded and $E$ is frozen. No gradient
from any control task ever reaches it, and every robot carries the same weights.

\subsection{Broadcast and anchor}

Every delivered message is prefixed with the sender's position relative to the receiver,
normalized by the communication radius and added at delivery time identically for every
condition. We introduced the anchor after finding that an egocentric latent is spatially
unusable without it. A neighbor's latent says ``hazard to my north-east'' but the
receiver cannot place ``my''. With the anchor in every condition, the experimental
question focuses on what the content adds beyond position.

\subsection{Receiver}

The receiver is a single cross-attention layer in the standard attention-fusion
form~\cite{das2019tarmac}. The navigator's own embedding is the query, the messages it
has received are the keys and values, and the pooled result is concatenated with the
ego embedding and passed to the policy. Pooling is permutation-invariant over neighbors,
and a neighbor that is out of range or absent is masked and contributes exactly zero.
The value projection is zero-initialized, so
every message condition starts as the no-message policy and recruits its channel only
when gradients justify it. Without this, a high-variance latent under a random
projection acts as a distractor that PPO does not escape under sparse reward.

\subsection{Route head and executor}
\label{sec:head}

Under continuous control the navigator's policy is decomposed into a decision and its
execution. A small route head reads the message and emits one categorical route per
episode. The route is committed as soon as the first message has arrived and is credited
with the whole episode return. A frozen, route-conditioned low-level controller then
drives the chosen route. The head has no ego input and the controller cannot read the
message, so the route decision has exactly one possible information source. In both
substrates the receiving module is trained by policy-gradient reinforcement learning
from the team reward~\cite{schulman2017ppo,yu2022mappo}, and ``reward-only'' refers to
that module and nothing else. Section~\ref{sec:port} explains why the decomposition is
needed under continuous control and not in the gridworld.

\section{From Grid Actions to Velocity Commands}
\label{sec:port}

Direct transfer of the gridworld protocol to continuous velocity control fails, and in
this section we analyze why. We first establish that the failure lies neither in
perception nor in communication, then show that the corridor decision is never sampled
under Gaussian exploration (Section~\ref{sec:never}), derive and validate a scalar
predictor of when this occurs (Section~\ref{sec:geometry}), show why increasing the
exploration noise does not help (Section~\ref{sec:harder}), and present the remedy
(Section~\ref{sec:remedy}).

The protocol of Section~\ref{sec:grid} transferred without change to rendered pixels of
the same arena (Fig.~\ref{fig:scene}b, Section~\ref{sec:setup}). Under velocity
control, six training configurations that varied the hazard penalty, the discount
factor, the spawn curriculum and the arena layout all converged to the same policy,
which commits to a single corridor regardless of slab placement (route optimality
${\approx}0.5$). Standard diagnostics did not localize the failure, since linear probes
recovered the hazard from the latent, the message arrived intact, and the controller
executed its commands. To rule out the perception and communication path entirely, we
replaced the learned latent with an oracle bit indicating the blocked corridor. This
condition also failed, with route optimality of $0.44$ and $0.53$ in two runs against
$0.51$ with no message ($n = 2$, indicative only). Since a noiseless message conveys the
decisive information exactly, its failure localizes the fault downstream of the channel,
in the policy's ability to act on what it receives.

\subsection{Exploration coverage of the corridor decision}
\label{sec:never}

\begin{table}[t]
\centering
\caption{Coverage (alternative-corridor entry) and task success under a swept
exploration distribution, $128$ episodes per row, one pretrained navigator, message
removed, no learning. $\sigma$ is the imposed lateral noise (the trained value is
$0.601$); $\tau$ is the AR(1) correlation time in control steps, $\tau = 0$ being iid.}
\label{tab:sweep}
\footnotesize
\begin{tabular}{@{}rrrr@{}}
\toprule
lateral $\sigma$ & $\tau$ & coverage & success \\
\midrule
0.55 & 0   & \textbf{0.00} & 0.68 \\
0.55 & 30  & \textbf{0.00} & 0.61 \\
1.00 & 0   & \textbf{0.00} & 0.58 \\
1.00 & 30  & 0.01 & 0.31 \\
1.00 & 100 & 0.02 & 0.22 \\
1.65 & 30  & 0.16 & 0.11 \\
1.65 & 100 & 0.09 & 0.09 \\
\bottomrule
\end{tabular}
\end{table}

To test whether the corridor decision is ever sampled, we take a pretrained navigator,
remove the message, and measure how often it enters the alternative corridor as a
function of the exploration noise (Table~\ref{tab:sweep}). At a lateral noise of
$\sigma = 0.55$, close to the trained value of $0.601$, the alternative corridor is
entered in $0$ of $128$ episodes. At $\sigma = 1.00$ the count remains $0$. Coverage
becomes nonzero only at noise amplitudes at which task success has already collapsed,
and across the sweep the two quantities move in opposite directions. The policy-gradient
update that would associate the message with the alternative corridor is an expectation
over trajectories that enter it, and that expectation is estimated from no samples.

The oracle-bit configuration shows what the channel is used for instead. Flipping the
bit shifts the policy's lateral mean by $2.69$ standard deviations, yet under
deterministic rollouts the policy enters the incumbent corridor in $100\%$ of episodes
for both slab placements and both bit values. The bit modulates progress \emph{within}
the corridor (success $1.00 \to 0.41$) but never the choice of corridor. Reward-driven
learning assigns the message whatever function the exploration distribution exposes.

\subsection{A scalar predictor of learnability}
\label{sec:geometry}

The measurement follows from the scene geometry. The navigator spawns equidistant from
the two corridor centers, $1.25$\,m laterally from each. With $v_{\max} = 0.5$\,m/s and
a $0.1$\,s control period, one step of full lateral command displaces the robot by
$0.05$\,m, so reaching the far corridor within an approach window of $N \approx 30$
steps requires a mean normalized lateral command of $d = 1.25/(0.05 \times 30) = 0.833$.
Under iid Gaussian exploration the window mean has standard deviation $\sigma/\sqrt{N}$,
so the required deviation lies
\begin{equation}
z = \frac{d\sqrt{N}}{\sigma} \approx 7.6
\label{eq:z}
\end{equation}
standard deviations from the mean at the trained $\sigma$, an event with probability of
order $10^{-14}$ per window. This is a lower bound. The trained policy holds a lateral
mean near $-d$ during its own approach, which places the alternative near $2d$ away, and
clamping to $[-1,1]$ truncates the tail further. In the gridworld one action moves a
whole cell, so the route is a single categorical draw and both routes appear in every
batch regardless of the current policy. The discrete action space supplies temporal
abstraction implicitly.

\begin{figure}[t]
\centering
\includegraphics[width=\columnwidth]{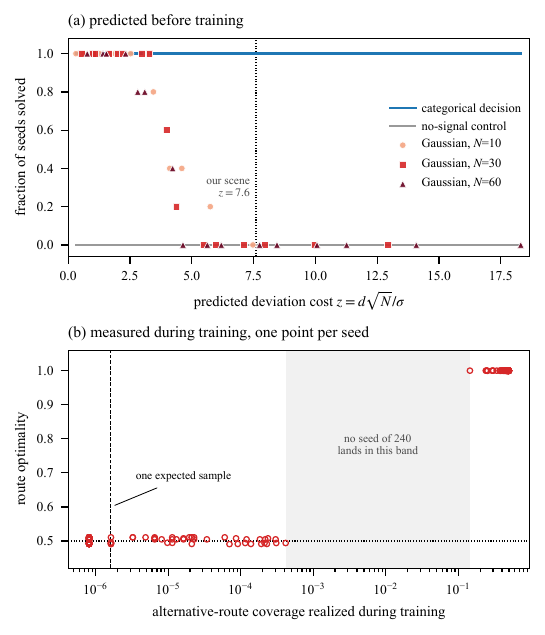}
\caption{The point-mass study. \textbf{(a)}~Fraction of seeds solved against
$z = d\sqrt{N}/\sigma$, computed before training from task geometry alone, five seeds per
setting; marker shape encodes window length $N$, and a sixfold range of $N$ collapses
onto one curve. The dotted line marks our scene. The categorical arm is flat at $1.0$;
the no-signal control never solves. \textbf{(b)}~The same runs against coverage realized
during training; outcomes are bimodal with complete separation.}
\label{fig:generality}
\end{figure}

We take $z$ as a predictor of learnability and validate it on a task that shares no
code, geometry or simulator with our scene (Fig.~\ref{fig:generality}). A
two-dimensional point mass in a corridor receives a one-bit message indicating the safe
side and must sustain a mean lateral command of at least $d$ over the first $N$ steps to
reach it. A Gaussian policy with a message-conditioned mean is trained by the
score-function gradient with Adam, alongside a no-signal control in which the safe side
is drawn independently of the message and a categorical arm that draws the side once per
episode. We sweep $N \in \{10, 30, 60\}$,
eight values of $d$ from $0.1$ to $1.3$ and $\sigma \in \{0.55, 1.0\}$, for $48$
settings, three arms, five seeds and $720$ runs. The fraction of seeds that solve the
task falls from $1.0$ to $0.0$ along a single sigmoid in $z$ centered near $4.2$, and
settings with equal $z$ but a sixfold difference in $N$ fall on the same curve. Outcomes
are bimodal. Of $240$ Gaussian seeds, $138$ finish at ceiling and $102$ at chance with
none between, and plotted against the coverage realized during training the two groups
separate completely. The categorical arm solves every setting up to $z = 18.3$, and the
no-signal control never exceeds $0.508$.

Two caveats apply. The optimizer is Adam because under SGD the threshold depends on the
learning rate (the majority-solved edge moves from $z \approx 3.1$ to $5.8$ as the rate
increases from $0.01$ to $1$). Adam's per-parameter normalization in turn allows the
policy mean to random-walk, which acts as a second exploration channel akin to
parameter-space noise~\cite{plappert2018paramnoise} and explains why the edge in $z$ is
soft while the edge in realized coverage is sharp.

\subsection{Effect of increased exploration noise}
\label{sec:harder}

Increasing $\sigma$ widens the distribution of the window mean, but the policy-gradient
estimator weights each action by the score function
$\partial \log \pi / \partial \mu = (a - \mu)/\sigma^2$, whose magnitude scales as
$1/\sigma$. The same $\sigma$ that makes a deviation reachable attenuates the gradient
that reports it by the same factor, and the cost appears in the success column of
Table~\ref{tab:sweep}. Temporally correlated noise is the standard
remedy~\cite{ruckstiess2008sde,raffin2021gsde,eberhard2023pink} and does reach the
decision. An AR(1) process with $\tau = 30$, confined to the first $40$ steps and to the
lateral axis, achieves $0.22$ coverage while success remains at $0.51$. Applied over the
whole episode, however, it destroys competence ($0.12$ coverage at $0.10$ success), and
confined to the window its signal is outweighed. The window occupies
$N_{\mathrm{in}} = 40$ of $600$ steps and per-sample gradient magnitude scales as
$1/\sigma$, so the aggregate gradient from the $560$ out-of-window steps exceeds the
in-window aggregate by a factor
$(N_{\mathrm{out}}/N_{\mathrm{in}})(\sigma_{\mathrm{in}}/\sigma_{\mathrm{out}})$, equal
to $14$ without a boost and growing linearly with it. Each out-of-window step reinforces
whatever route was executed there, which in most episodes is the incumbent. Empirically,
the alternative-corridor success obtained with the boost held at $0.19$--$0.26$ through
iteration $28$ and had decayed to $0.00$--$0.06$ by iteration $80$, while the boost was
still $3.6\times$ the policy's own noise. The boost anneals over this period, so a
single run cannot distinguish active suppression from decay, and the ratio above, which
treats in- and out-of-window advantages as equal in magnitude, is a heuristic rather
than a proof. We see no reason to expect the mechanism to be specific to this scene.

\subsection{Restoring temporal abstraction}
\label{sec:remedy}

Since the discrete action space supplied temporal abstraction implicitly, we restore it
explicitly for the one decision that requires it. The route is drawn once per episode
from a categorical head (Section~\ref{sec:head}), which sets $N = 1$ for that decision
while leaving $\sigma$ unchanged. The head's logits are zero-initialized, so it starts
uniform and every batch contains both routes in equal proportion, a sampling structure
that none of the six failed configurations had. The construction is an option in the
sense of Sutton \emph{et al.}~\cite{sutton1999options}; the contribution is not the
mechanism but the observation that its absence silently breaks the transfer to
continuous control. Consistent with the analysis, redrawing the decision
every $30$ steps or every step degrades route optimality to $0.920$ and $0.768$
respectively (Table~\ref{tab:race}, lower block), monotonically in the redraw rate.

\section{Experiments}
\label{sec:experiments}

We design the evaluation to answer four questions. \textbf{(Q1)}~Does perceptual
content separate from kinematic content when only the content varies?
\textbf{(Q2)}~Does a trained policy actually read the message at decision time?
\textbf{(Q3)}~Does the result survive rendered pixels and continuous control?
\textbf{(Q4)}~Does the encoder survive a real camera?

\subsection{Setup}
\label{sec:setup}

\textbf{Task.} Two holonomic robots, a walled arena split by a barrier pierced by two
corridors, and a hazard slab placed uniformly at random in one corridor each episode
(Fig.~\ref{fig:scene}b). The navigator must reach a goal on the far side, and crossing
the slab is penalized. The scout is stationary and can see which corridor is blocked.
The navigator cannot. We verify the occlusion rather than assume it. At both
corridor-choice points the navigator's camera contains zero hazard pixels under both
slab placements (Fig.~\ref{fig:scene}c), and an earlier layout that leaked ten pixels
is reproducible with a flag. The same layout is instantiated twice. The gridworld has
$0.5$\,m cells, a shadow-cast field of view and five discrete actions. The rendered
scene runs in a GPU-accelerated physics simulator with physics at $60$\,Hz, control at
$10$\,Hz, a $64{\times}64$ RGB forward camera, body-frame velocity commands
$(v_x, v_y, \omega) \in [-1,1]^3$ scaled by $0.5$\,m/s and $1.5$\,rad/s, and
$600$-step episodes.

\begin{table}[t]
\centering
\caption{Message contents compared. Every row shares the anchor, delivery, latency,
receiver and training budget; only the content differs. Widths are for the pixel
experiments (in the gridworld the raw observation is a $225$-cell egocentric patch).}
\label{tab:conditions}
\scriptsize\setlength{\tabcolsep}{3pt}
\begin{tabular}{@{}lllr@{}}
\toprule
role & condition & content $f(o^i_t)$ & width \\
\midrule
lower bound & floor & anchor + zeros & 66 \\
kinematic & position & anchor + $(x,y)$, zero-padded & 66 \\
kinematic (strong) & trajectory & anchor + $(x,y)$ + const.-vel.\ path & 66 \\
ours (C1) & latent \zt{} & anchor + $E(o^i_t)$ & 66 \\
ours (C2) & predicted \zhat{} & anchor + $P(E(o^i_t), a)$ & 66 \\
upper bound & raw observation & anchor + $64{\times}64{\times}3$ frame & 12,290 \\
diagnostic & oracle & anchor + true hazard bit, zero-padded & 66 \\
\bottomrule
\end{tabular}
\end{table}

\textbf{Conditions and metrics.} Table~\ref{tab:conditions} instantiates the protocol of
Section~\ref{sec:formulation}. The latent is $64$-dimensional and the anchor two floats,
so the wire is $66$ floats wide for every condition except the raw image, which is $186$
times wider. \emph{Route optimality} is the fraction of episodes in which the corridor
entered, or committed to, is the slab-free one. \emph{Task success} is the fraction
reaching the goal and \emph{hazard contacts} the number of slab-crossing steps per
episode. In Section~\ref{sec:port}, \emph{coverage} is the fraction of episodes entering
the non-incumbent corridor under a given exploration distribution.

\textbf{Implementation.} The encoder has three convolutions ($32$, $64$, $64$ channels,
kernels $8/4/3$, strides $4/2/1$) and a linear map to $64$ dimensions, trained on
$204{,}800$ rendered frames from random rollouts with uniform free-pose spawns. The
freeze gates it passed were a linear hazard-visible probe at $93.5\%$ against an $85.7\%$
majority baseline, wall-count $R^2$ of $0.985$ and effective rank $44.5$ of $64$. In the
gridworld the receiver is trained with MAPPO~\cite{yu2022mappo}, a shared-parameter actor
and a centralized critic that sees the ground-truth hazard placement, with the scout
masked from the update, for $1.2$M environment steps per seed. In the pixel experiments
the route head has $4{,}483$ parameters (one hidden layer of $64$ units feeding two
logits and a value), commits at step~$2$, and trains at learning rate
$3{\times}10^{-3}$ on batches of $256$ episodes for $6{,}000$ episodes across $64$
parallel environments. The controller is pretrained beforehand with route-conditioned
potential-based shaping~\cite{ng1999shaping} and an obedience-as-success reward at
$2.5{\times}10^{-4}$, then frozen. It enters the commanded corridor in $1.000$ of
episodes (\emph{obedience}). PPO settings are shared everywhere: clip $0.2$, four
epochs, entropy coefficient $0.01$, $\gamma = 0.99$, $\lambda = 0.95$, the same for
every seed and condition.

\subsection{Gridworld (Q1)}
\label{sec:grid}

Six conditions, five seeds each. The frozen latent \zt{} solves the task in $5$ of $5$
seeds (success $1.00 \pm 0.00$), the predicted latent \zhat{} in $4$ of $5$, and the raw
$225$-cell observation in $3$ of $5$. Floor, position and trajectory solve $0$ of $15$
and remain at the $0.47$ chance level. The $64$-dimensional latent thus solves more
seeds than the $225$-dimensional patch it was computed from ($5$ versus $3$ of $5$),
although at this $n$ the difference is suggestive rather than significant. Against the
floor the latent has no seed overlap (exact one-sided rank test, $p = 0.004$, the
minimum attainable at $n = 5$). Because the scout is stationary, its position and
trajectory are constant within an episode and carry no information about the hazard, so
the comparison measures what perceptual content adds beyond position at matched
bandwidth.

\subsection{Rendered pixels under continuous control (Q1, Q3)}
\label{sec:pixels}

\begin{table}[t]
\centering
\caption{Pixel experiments. Route optimality and task success over the last $500$ of
$6{,}000$ training episodes, the head's \emph{sampled} choice, mean $\pm$ s.d.\ over
seeds; executor obedience is $1.000$ throughout. Lower block: the same head, wire and
executor with the route redrawn every $k$ steps (Section~\ref{sec:port}).}
\label{tab:race}
\footnotesize\setlength{\tabcolsep}{4pt}
\begin{tabular}{@{}lccc@{}}
\toprule
message content & $n$ & route opt. & task success \\
\midrule
latent \zt{}                   & 5 & $\mathbf{0.997 \pm 0.006}$ & $0.992 \pm 0.007$ \\
predicted \zhat{}              & 3 & $0.995 \pm 0.005$ & $0.991 \pm 0.006$ \\
oracle (noiseless bit)         & 5 & $0.988 \pm 0.006$ & $0.984 \pm 0.003$ \\
raw observation ($12{,}290$-d) & 3 & $0.819 \pm 0.230$ & $0.815 \pm 0.228$ \\
\midrule
floor (anchor + zeros)         & 5 & $0.496 \pm 0.020$ & $0.496 \pm 0.021$ \\
floor, whole wire zeroed       & 3 & $0.500 \pm 0.016$ & $0.499 \pm 0.015$ \\
position                       & 3 & $0.484 \pm 0.005$ & $0.482 \pm 0.007$ \\
trajectory                     & 3 & $0.477 \pm 0.018$ & $0.476 \pm 0.017$ \\
\midrule
\zt{}, one draw ($k{=}0$)      & 5 & $0.994 \pm 0.008$ & $0.988 \pm 0.013$ \\
\zt{}, every $30$ steps        & 5 & $0.920 \pm 0.021$ & $0.915 \pm 0.020$ \\
\zt{}, every step ($k{=}1$)    & 5 & $0.768 \pm 0.049$ & $0.770 \pm 0.048$ \\
oracle, every step ($k{=}1$)   & 5 & $0.929 \pm 0.022$ & $0.926 \pm 0.023$ \\
\bottomrule
\end{tabular}
\end{table}

\begin{figure*}[t]
\centering
\includegraphics[width=\textwidth]{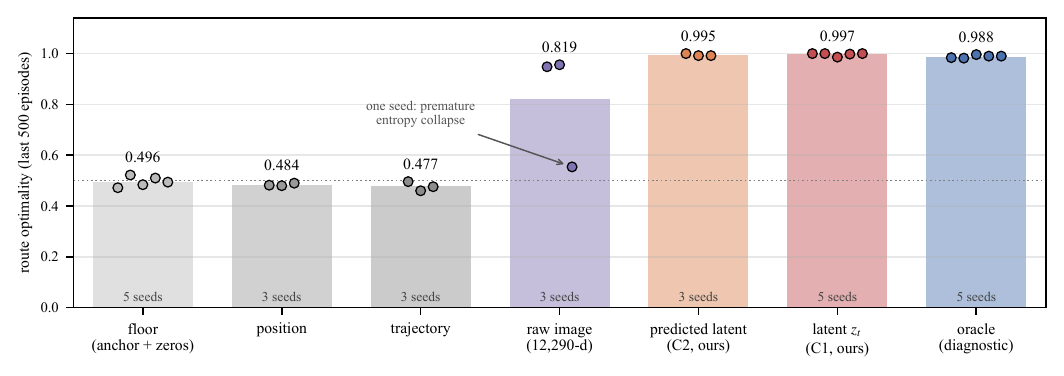}
\caption{Route optimality on pixels under continuous velocity control, seven message
contents, per-seed dots, dotted line at chance. Position and trajectory sit on the floor
with the empty wire; the $66$-float latent matches the noiseless oracle; the raw image,
$186\times$ wider, loses a seed to premature entropy collapse.}
\label{fig:sweep}
\end{figure*}

The same protocol on rendered pixels, with the frozen executor and route head, gives
Table~\ref{tab:race} and Fig.~\ref{fig:sweep}. The latent reaches $0.997 \pm 0.006$
route optimality against an anchored floor of $0.496 \pm 0.020$, with no seed overlap
($0.986$--$1.000$ versus $0.472$--$0.522$, $p = 0.004$), and matches the noiseless
oracle ($0.988 \pm 0.006$). The table scores the sampled choice and therefore includes
residual entropy; evaluated greedily, both reach $1.000$ (Table~\ref{tab:corrupt},
intact row). Task success tracks route optimality within $0.005$ throughout, hazard
contacts are zero for the latent and the oracle, and the predicted latent performs
identically. Position and trajectory reach $0.484$ and $0.477$, indistinguishable from
the floor. The anchored and zero-wire floors agree, confirming that the anchor carries
no route information for a stationary scout. The raw image, $186\times$ wider, reaches
$0.819 \pm 0.230$. Two seeds finish near $0.95$ and one collapsed its decision entropy
to ${\sim}10^{-6}$ and remained at chance. With three seeds against five the comparison
is not significant (Fisher $p = 0.375$), but its direction is opposite to what a
bandwidth argument alone would predict. The compressed latent is not only feasible for
a radio link but also easier for the policy to exploit under sparse reward.

\subsection{Corruption controls (Q2)}

\begin{table}[t]
\centering
\caption{Frozen trained heads re-evaluated greedily under wire corruption, $256$
episodes per cell: route optimality of three \zt{} heads (mean and range) and of the
oracle head.}
\label{tab:corrupt}
\footnotesize
\begin{tabular}{@{}lccc@{}}
\toprule
wire at evaluation & \zt{} mean & \zt{} range & oracle \\
\midrule
intact                 & $\mathbf{1.000}$ & $1.000$--$1.000$ & $1.000$ \\
content zeroed         & $0.507$ & $0.496$--$0.527$ & $0.547$ \\
whole wire zeroed      & $0.520$ & $0.496$--$0.555$ & $0.488$ \\
sender shuffled        & $0.492$ & $0.445$--$0.520$ & $0.547$ \\
matched-scale noise    & $0.458$ & $0.410$--$0.484$ & $0.555$ \\
\bottomrule
\end{tabular}
\end{table}

Recruitment is a causal claim, so we test it by destroying the message content of a
trained policy with everything else held fixed. Each trained head is frozen and
re-evaluated greedily under five wire conditions (Table~\ref{tab:corrupt}). With the
wire intact every head scores $1.000$. Zeroing the content, zeroing the whole wire,
substituting a real message from another environment, or replacing the content with
matched-scale Gaussian noise each return the same head to $0.41$--$0.56$, the oracle
head included. Under zeroed content the head collapses to a constant corridor and
episode length rises from ${\sim}209$ to $380$--$440$ steps. The route is therefore
carried by the content alone. Nor does the reward provide indirect supervision. The head
is never told whether its corridor matched the bit; errors are penalized only through
slab contact and successes rewarded only through arrival. In the gridworld the same
corruption collapses $10$ of the $12$ successful policies. The two exceptions learned a
message-free peek-and-reroute strategy, visible in their episode length, and the
never-solved checkpoints are invariant to corruption, which serves as a negative control.

\subsection{Hardware (Q4)}
\label{sec:hardware}

Since the head consumes latents, the message path meets the reality gap only at the
encoder, and we test that module on a physical robot (Fig.~\ref{fig:scene}a). The
physical camera's central $64{\times}64$ crop subtends $32^{\circ}$ (tape-verified),
against the $82.3^{\circ}$ pinhole of the simulation results above, so the encoder was
rebuilt at matched optics, retrained with segmentation-guided appearance randomization
of floor and walls, selected among four candidates by pre-registered gates, and frozen.
A $65$-parameter linear hazard probe fitted on simulated latents transfers its
\emph{direction} to real frames (AUC $0.964$) but not its operating point (balanced
accuracy $0.500$), so the pre-registered fallback refit the probe on real latents from
one session ($53$ frames). On a quarantined validation session at unseen distances under
dimmer light it scores $0.929$ balanced accuracy. Live, over $34$ fresh placements
at $120$--$250$\,cm under both lighting states, it decoded $102$ of $102$ decisions
($57$ hazard, $45$ clear) correctly, for $1.000$ balanced accuracy against a
pre-registered gate of $0.95$ and a Wilson $95\%$ lower bound of $0.964$ over decisions
and $0.90$ over placements. A raw-pixel red-fraction rule also scores $1.000$ on these
frames; the purpose of the test is to establish that a frozen encoder retains a bit it
was never trained to preserve, which it does. The robot was not driven closed-loop. The
matched-optics executor is bang-bang ($84$--$97\%$ of pre-clamp action components past
the rails on held-out simulated frames) and a static bench cannot certify it safe. This
is a decode-level test of the one module on which the message depends.

\section{Related Work}
\label{sec:related}

\textbf{Learned communication in MARL.} Differentiable and reinforced channels learn the
message end-to-end from task
reward~\cite{foerster2016dial,sukhbaatar2016commnet,das2019tarmac,singh2019ic3net}; Lin
\emph{et al.} ground the message in an autoencoder trained jointly with the
policy~\cite{lin2021ground}. Intention sharing broadcasts imagined
trajectories~\cite{kim2021intention}, and learned multi-agent path finding exchanges
positions and local plans~\cite{sartoretti2019primal,ma2021dhc}. These methods learn the
message jointly with the policy or fix it to kinematics; we fix the message before any
policy exists, as a perceptual latent, and learn only the readout, from reward.

\textbf{Collaborative perception.} Who2com, When2com, V2VNet, Where2comm and the
multi-robot GNN perception of Zhou \emph{et al.} share intermediate features between
agents under bandwidth
constraints~\cite{liu2020who2com,liu2020when2com,wang2020v2vnet,hu2022where2comm,zhou2022gnn}.
They supervise fusion against a perception label; we supervise nothing but the task, and
the shared feature is task-agnostic.

\textbf{Self-supervised world models and broadcast latents.} Joint-embedding predictive
encoders and world models produce, without labels, the perceptual latent a teammate
would need~\cite{assran2023ijepa,assran2025vjepa2,zhou2024dinowm}. Concurrent CS-JEPA
and Social-JEPA broadcast predictive latents between agents and read them with ridge
probes on labeled episodes~\cite{gazzaev2026csjepa,socialjepa2026}. We share the
architecture with this line and differ in the readout: a control decision trained from
scalar reward, between asymmetric agents, on continuous action spaces, with a hardware
test.

\textbf{Exploration and temporal abstraction in continuous control.} State-dependent,
colored and parameter-space noise~\cite{ruckstiess2008sde,raffin2021gsde,eberhard2023pink,plappert2018paramnoise},
options~\cite{sutton1999options}, action
persistence~\cite{metelli2020persistence,biedenkapp2021temporl} and bang-bang
policies~\cite{seyde2021bangbang} all address the weakness of iid Gaussian exploration.
We reach the same diagnosis and add that, for a communicated decision, restoring the
abstraction for that one decision is sufficient.

\section{Limitations, Conclusion and Future Work}
\label{sec:conclusion}

\textbf{Limitations.} One scene, one map seed and one encoder seed per substrate. The
pixel result is a contextual bandit over a frozen executor, not end-to-end multi-agent
learning, and the executor's route-conditioned pretraining is part of the remedy rather
than a neutral substrate. The decisive fact is one bit and a hand-designed oracle ties
the latent, so we have shown that a frozen latent is \emph{sufficient} for reward-only
recruitment, not that it is necessary. The scout is stationary. The hardware test is
decode-only.

\textbf{Conclusion.} We proposed Latent Telepathy, in which decentralized robots
broadcast the frozen self-supervised latent they already compute, and showed that a
teammate can learn to act on it from task reward alone, at parity with a noiseless
hand-designed message, on
pixels, under continuous control, and through a real camera. Because the encoder already
runs for perception, the scheme costs nothing extra to compute and a few dozen floats to
send. The diagnostic that emerged along the way is inexpensive to apply. If a noiseless
oracle message does not improve the policy, the fault lies in the action space rather
than in the channel.

\textbf{Future work.} Each limitation corresponds to an experiment. A richer decision
(variable hazard depth, or a choice among several corridors) would test whether the
latent is necessary rather than sufficient, and is the one we consider most valuable. A
mobile scout would make position and trajectory informative and allow a fair comparison
against them. Closed-loop driving on hardware would test the controller rather than the
encoder.

\bibliographystyle{IEEEtran}
\bibliography{refs}

\end{document}